\documentclass[conference]{IEEEtran}
\IEEEoverridecommandlockouts
\usepackage{cite}
\usepackage{amsmath,amssymb,amsfonts,caption}
\usepackage{graphicx}
\usepackage{textcomp}
\usepackage{xcolor}
\usepackage{multirow}
\usepackage[usestackEOL]{stackengine}

\usepackage[ruled,vlined,linesnumbered]{algorithm2e}

\def\BibTeX{{\rm B\kern-.05em{\sc i\kern-.025em b}\kern-.08em
    T\kern-.1667em\lower.7ex\hbox{E}\kern-.125emX}}

\newcommand{\etal}{\textit{et al}.}

\begin{document}

\title{Zero-Shot Anomaly Detection in Battery Thermal Images Using Visual Question Answering with Prior Knowledge\\
\thanks{The research work presented in this article is financially supported by the European Union’s Horizon Europe project ENERGETIC (Grant No 101103667).}
}

\author{\IEEEauthorblockN{Marcella Astrid \qquad
Abdelrahman Shabayek \qquad Djamila Aouada}
\IEEEauthorblockA{Interdisciplinary Centre for Security, Reliability and Trust (SnT), University of Luxembourg, Luxembourg \\
Email: marcella.astrid@uni.lu,
abdelrahman.shabayek@uni.lu,
djamila.aouada@uni.lu}}

\maketitle

\begin{abstract}
Batteries are essential for various applications, including electric vehicles and renewable energy storage, making safety and efficiency critical concerns. Anomaly detection in battery thermal images helps identify failures early, but traditional deep learning methods require extensive labeled data, which is difficult to obtain, especially for anomalies due to safety risks and high data collection costs. To overcome this, we explore zero-shot anomaly detection using Visual Question Answering (VQA) models, which leverage pretrained knowledge and text-based prompts to generalize across vision tasks. By incorporating prior knowledge of normal battery thermal behavior, we design prompts to detect anomalies without battery-specific training data. We evaluate three VQA models (ChatGPT-4o, LLaVa-13b, and BLIP-2) analyzing their robustness to prompt variations, repeated trials, and qualitative outputs. Despite the lack of fine-tuning on battery data, our approach demonstrates competitive performance compared to state-of-the-art models that are trained with the battery data. Our findings highlight the potential of VQA-based zero-shot learning for battery anomaly detection and suggest future directions for improving its effectiveness.
\end{abstract}


\section{Introduction}

Batteries have become an essential part of modern technology, supporting a wide range of applications. In the automotive industry, they serve as the core energy source for electric vehicles, replacing traditional gasoline-powered engines known for their adverse effects on human health and the environment \cite{peters2020public}. In renewable energy systems, batteries store electricity generated from wind and solar power \cite{lehtola2019solar}. Given their widespread use, ensuring battery safety is crucial. Anomaly detection plays a key role in the early identification of potential battery failures \cite{wu2023anomaly} while also contributing to efficiency \cite{haider2020data}. Since temperature is one of the key parameters in battery monitoring \cite{wang2019wireless}, our work focuses on anomaly detection in battery thermal images.

Anomaly detection with deep learning has been widely used in various vision applications \cite{hyun2024reconpatch,astrid2024exploiting}. However, collecting data, especially anomalous data, can be challenging, particularly in battery applications where it may pose safety risks. As a result, supervised learning methods that require anomalous data are difficult to implement. To address this, recent methods have focused on training with only normal data \cite{shabayek2025ai}. However, even collecting normal/non-anomalous data is expensive \cite{herle2021overcoming,liu2024status}, such as the time cost of cycling through the battery charge and discharge phases. Therefore, we aim to explore whether zero-shot learning can be a viable alternative to detect anomalies without any battery training data.

\begin{figure*}[]
  \centering
  \includegraphics[width=\linewidth]{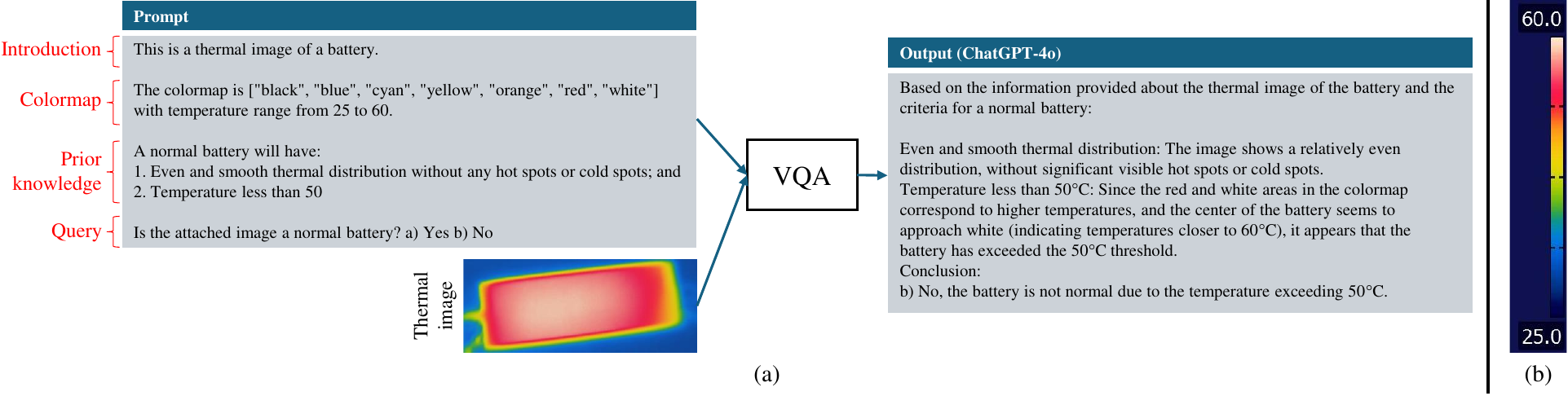}
  \vspace{-5mm}
  \caption{(a) Our proposed zero-shot anomaly detection method takes a text prompt and a thermal image as input. The text prompt includes information about the type of image provided (introduction), the colormap range and color, prior knowledge of normal battery characteristics, and the query. (b) Illustration of the colormap used in the thermal image.}
  \vspace{-2mm}
  \label{fig:method}
\end{figure*}

Visual Question Answering (VQA) pretrained with a large amount of data has been utilized for zero-shot learning in many vision applications due to its strong generalization capability \cite{guo2023images,zhang2024gpt,castrillon2024visual}. It uses text prompts to perform tasks in vision-based applications. For our purpose, we have prior knowledge of what a normal battery should look like. As mentioned in \cite{shabayek2025ai}, a normal battery should have a homogeneous thermal distribution and should not overheat (i.e., the temperature stays below a certain threshold). Therefore, we design a prompt exploiting this information as a context to detect anomalies in a zero-shot manner, as illustrated in Figure \ref{fig:method}(a).

In this work, we explore three VQA models (ChatGPT-4o \cite{hurst2024gpt}, LLaVa-13b\footnote{https://replicate.com/yorickvp/llava-13b} \cite{liu2024improved}, and BLIP-2\footnote{https://replicate.com/andreasjansson/blip-2} \cite{li2023blip}) for zero-shot anomaly detection in battery thermal images. We analyze their variability across different rephrased prompts, repeated trials, and qualitative behaviors to identify their strengths and weaknesses. Our simple yet effective approach is compared with state-of-the-art (SOTA) models specifically trained on battery data, and we find that our method remains competitive despite not being fine-tuned on battery data. Finally, we conclude with key findings and discuss potential future research directions based on our observations.

\begin{table*}[]
\centering
\caption{Five different prompts are used in this work. The first prompt is handcrafted, while the second to fifth prompts are rephrased using ChatGPT-4o based on the first prompt.}
\vspace{-1mm}
\resizebox{\linewidth}{!}{
\begin{tabular}{|l|l|l|l|l|}
\hline
\multicolumn{1}{|c|}{Prompt 1} & \multicolumn{1}{c|}{Prompt 2} & \multicolumn{1}{c|}{Prompt 3} & \multicolumn{1}{c|}{Prompt 4} & \multicolumn{1}{c|}{Prompt 5} 
\tabularnewline \hline
\begin{minipage}[t]{0.2\linewidth}%
This is a thermal image of a battery. \\ \\ The colormap is {[}"black", "blue", "cyan", "yellow", "orange", "red", "white"{]} with temperature range from 25 to 60. \\ \\ A normal battery will have:\\ 1. Even and smooth thermal distribution without any hot spots or cold spots; and\\ 2. Temperature less than 50\\ \\ Is the attached image a normal battery? a) Yes b) No
\end{minipage} &  
\begin{minipage}[t]{0.2\linewidth}%
This is a thermal image of a battery.\\ \\ The color scale used is {[}"black", "blue", "cyan", "yellow", "orange", "red", "white"{]}, representing a temperature range of 25 to 60°C.\\ \\ A normal battery should exhibit:\\ 1. A smooth and even thermal distribution without any hot or cold spots.\\ 2. A maximum temperature below 50°C.\\ \\ Does the attached image indicate that the battery is normal?\\ a) Yes\\ b) No
\end{minipage} & 
\begin{minipage}[t]{0.2\linewidth}%
This thermal image shows a battery.\\ \\ The colormap used includes {[}"black", "blue", "cyan", "yellow", "orange", "red", "white"{]}, representing temperatures ranging from 25°C to 60°C.\\ \\ A normal battery should exhibit:\\ 1. An even and smooth thermal distribution, free of hot or cold spots; and\\ 2. A temperature below 50°C.\\ \\ Based on the attached image, is this battery normal?  \\ a) Yes  \\ b) No
\end{minipage} & 
\begin{minipage}[t]{0.2\linewidth}%
The attached thermal image displays a battery with the following colormap: {[}"black," "blue," "cyan," "yellow," "orange," "red," "white"{]}, corresponding to a temperature range of 25 to 60. \\ \\ A normal battery should meet these criteria: \\ 1. A smooth and even thermal distribution with no hot spots or cold spots. \\ 2. A temperature below 50. \\ \\ Based on this information, is the battery in the image normal?  \\ a) Yes  \\ b) No
\end{minipage} & 
\begin{minipage}[t]{0.2\linewidth}%
This thermal image represents a battery.\\ \\ The colormap ranges from "black" to "white" ({[}"black", "blue", "cyan", "yellow", "orange", "red", "white"{]}) corresponding to a temperature range of 25°C to 60°C.\\ \\ For a battery to be considered normal:\\ 1. It should exhibit an even and smooth thermal distribution without any hot or cold spots.\\ 2. The temperature should remain below 50°C.\\ \\ Based on these criteria, does the attached image show a normal battery?  \\ a) Yes  \\ b) No
\end{minipage} 
\tabularnewline \hline
\end{tabular}
}
\vspace{-4mm}
\label{tab:prompt}
\end{table*}

\noindent\textbf{Paper organization:} We discuss related work in Section~\ref{sec:related_work}. The method is detailed in Section~\ref{sec:methodology}. Section~\ref{sec:experiments} covers the experimental setup and results. Finally, our conclusion and future work are presented in Section~\ref{sec:conclusion}.

\section{Related Work}
\label{sec:related_work}

The two most related works to our study are Zhang \etal~on zero-shot vision anomaly detection using VQA \cite{zhang2024gpt} and Shabayek \etal~on anomaly detection in battery thermal images \cite{shabayek2025ai}. Zhang \etal~focus on industrial anomaly detection, assuming the presence of anomalous regions that differ from their surroundings. Their approach utilizes super-pixels to define these regions, which is particularly effective for detecting scratches or defects in real-world photographic images. However, due to the gradient nature of thermal images of batteries, this method may not be as effective for our purpose. Additionally, overheating anomalies without distinct hot or cold spots do not conform to the assumption of detectable regional differences. Instead, we leverage prior knowledge of normal thermal image characteristics to identify anomalies.

Shabayek \etal~use normal thermal images to train a model with pseudo-anomaly feature augmentation, simulating artificial defects. However, their method still requires training data, which can be costly and time-consuming to collect. To overcome this limitation, we propose a zero-shot approach that eliminates the need for training data while still enabling effective anomaly detection.



 \begin{figure*}[]
  \centering
  \includegraphics[width=\linewidth]{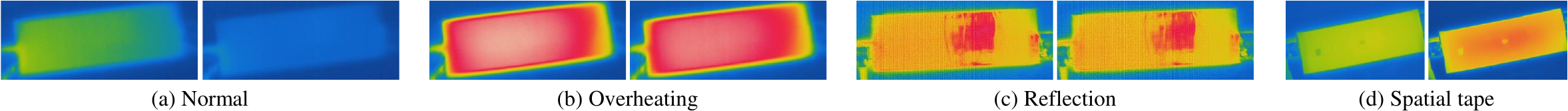}
  \vspace{-6mm}
  \caption{Samples from the test set proposed in \cite{shabayek2025ai}. It consists of normal images and three types of anomalies: overheating, reflection, and spatial tape. (a) Normal images show a smooth gradient and a temperature below the threshold. (b) Overheating images exhibit high overall temperatures, even without distinct hot or cold spots. (c) Reflection images display an uneven distribution with hot spots and abnormally high temperatures. (d) Spatial tape cases show cold spots.}
  \label{fig:test_set}
\end{figure*}

\begin{table*}[]
\centering
\caption{Accuracy (Acc.) (\%) averaged (Avg.) over multiple trials, range (max - min) (\%) of accuracy across multiple trials, and the percentage of unsure predictions across all trials. Tested on different test set splits with various prompts and VQA models. The number of trials is 5, 3, and 3 for ChatGPT-4o, LLaVa-13b, and BLIP-2, respectively.}
\vspace{-1mm}
\resizebox{\linewidth}{!}{
\begin{tabular}{|l||ccccc|ccccc|ccccc|}
\hline
Model              & \multicolumn{5}{c|}{ChatGPT-4o \cite{hurst2024gpt}}    & \multicolumn{5}{c|}{LLaVa-13b \cite{liu2024improved}}    & \multicolumn{5}{c|}{BLIP-2 \cite{li2023blip}}           \\ \hline
Prompt             & 1    & 2     & 3     & 4    & 5    & 1    & 2     & 3    & 4    & 5    & 1     & 2     & 3     & 4     & 5     \\ \hline \hline
Avg. Acc. (all)          & 73.0  & 82.3  & 71.0  & 79.7 & 82.3 & 58.3 & 55.6 & 68.9 & 63.3 & 52.2 & 55.0  & 55.0  & 55.0  & 55.0  & 88.3  \\ \hline
Avg. Acc. (normal)     & 42.2  & 63.0  & 37.8  & 60.0 & 75.6 & 11.1 & 2.5 & 93.8 & 29.6 & 6.2 &  0 & 0  & 0  & 0  &  74.1  \\
Avg. Acc. (anomaly)     & 98.2 & 98.2  & 98.2  & 95.8 & 87.9 & 97.0 & 99.0 & 48.5 & 90.9 & 89.9 &  100.0 & 100.0  & 100.0  & 100.0  &  100.0 \\ \hline
Avg. Acc. (overheating)             & 98.5  & 100.0  & 100.0  & 98.5 & 100.0  & 92.3 & 97.4 & 10.3  & 87.2  & 97.4 &  100.0 &  100.0 & 100.0  & 100.0  & 100.0   \\
Avg. Acc. (reflection)      & 100.0 & 100.0  & 100.0  & 100.0 & 100.0 & 100.0 & 100.0 & 100.0 & 100.0 & 86.1  & 100.0  &  100.0 &  100.0 & 100.0  & 100.0    \\
Avg. Acc. (spatial tape)      & 95.0 & 92.5  & 92.5  & 85.0 & 50.0 & 100.0 & 100.0 & 33.3 & 83.3 & 83.3 & 100.0  & 100.0  &  100.0 &  100.0 & 100.0  \\ \hline \hline
Range Acc. (all)          & 5.0  & 5.0  & 16.7  & 6.7 & 5.0 & 3.3  & 1.7 & 10.0 & 3.3 & 10.0  & 0.0  & 0.0  & 0.0  & 0.0  &  0.0 \\ \hline \hline
\%Unsure (all)          &  1.0 & 0.0 & 0.7  & 2.7 & 0.0 &  9.4 & 1.7 & 0.0 & 41.7 & 0.0  & 0.0  & 0.0  & 0.0  & 0.0  &  0.0 \\ \hline
\end{tabular}
}
\vspace{-3mm}
\label{tab:results}
\end{table*}

 \begin{figure*}[]
  \centering
  \includegraphics[width=\linewidth]{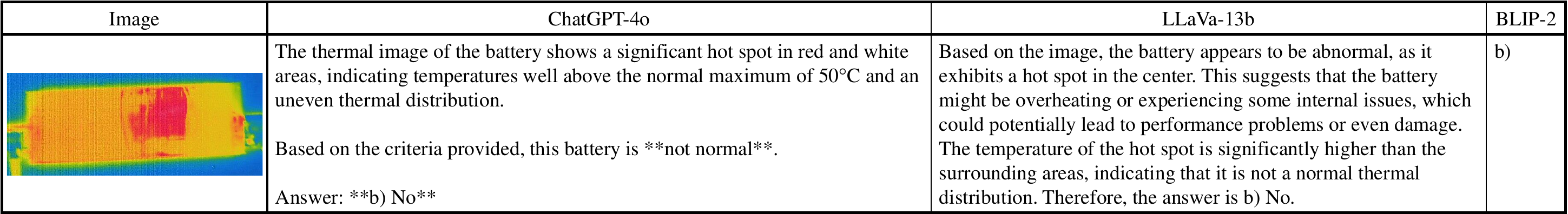}
  \vspace{-5mm}
  \caption{Output examples for normal and anomalous data from the three VQA models: ChatGPT-4o \cite{hurst2024gpt} (Prompt 2), LLaVa-13b \cite{liu2024improved} (Prompt 3), and BLIP-2 \cite{li2023blip} (Prompt 5). ChatGPT-4o and LLaVa-13b provide explanations in addition to their predictions, while BLIP-2 only generates the final prediction.}
  \vspace{-5mm}
  \label{fig:output_examples}
\end{figure*}

\section{Methodology}
\label{sec:methodology}

The overall method is illustrated in Figure~\ref{fig:method}(a). The VQA model takes a text prompt and a thermal image as input. Adding context to the prompt has been shown to improve VQA performance in previous works~\cite{hu2024multi,zhang2024gpt}. In our approach, the context includes an introduction to the type of image, the colormap used in the thermal image (as illustrated in Figure~\ref{fig:method}(b)), and prior knowledge of the normal thermal pattern. The prior knowledge specifies two key characteristics of normal images, as defined in \cite{shabayek2025ai}: a temperature below the 50 degree Celsius threshold and a smooth thermal distribution without distinct hot or cold spots. The query is the question related to anomaly detection. Since we utilize prior knowledge of normal data, we formulate the query to ask whether the image is normal or not.

As different prompts may generate different results \cite{wahle2024paraphrase}, we create four more prompts rephrased from our first handcrafted prompts. Table~\ref{tab:prompt} shows all prompts we use in this method. Prompt 1 is our handcrafted prompt. Prompt 2 to 5 are rephrased by ChatGPT-4o from Prompt 1. 

This process is performed using a VQA model \cite{hurst2024gpt,liu2024improved,li2023blip} pretrained on a large amount of generic data available online. We do not train or fine-tune the model with battery thermal images, making this a purely zero-shot approach. Despite its simplicity, our method remains effective, demonstrating competitive performance compared to specialized models trained on battery data.

\section{Experiments}
\label{sec:experiments}
\subsection{Dataset}


We use the real battery thermal image dataset from \cite{shabayek2025ai}, evaluating our zero-shot method solely on the test set (Figure~\ref{fig:test_set}). The dataset comprises four subsets: (a) normal, (b) overheating, (c) reflection, and (d) spatial tape.

Due to safety concerns, anomalies in (b)–(d) are artificially generated. Overheating images simulate high temperatures ($>$50°C) with a smooth distribution. Reflection images come from unpainted batteries, creating hot spots and uneven temperatures. Spatial tape images simulate cold spots on unpainted areas. The dataset includes 27, 13, 12, and 8 images for normal, overheating, reflection, and spatial tape, respectively.

\subsection{Experiment setup}
For each image and text prompt pair, we repeat the experiment five times for ChatGPT-4o and three times for the other models. The temperature for the LLaVa-13b model is set to 0.1. Unsure predictions can occur in a few cases (Section~\ref{subsubsec:unsure} for further details) and are considered anomaly class predictions, as this approach makes more sense in terms of safety.

 \begin{figure*}[]
  \centering
  \includegraphics[width=\linewidth]{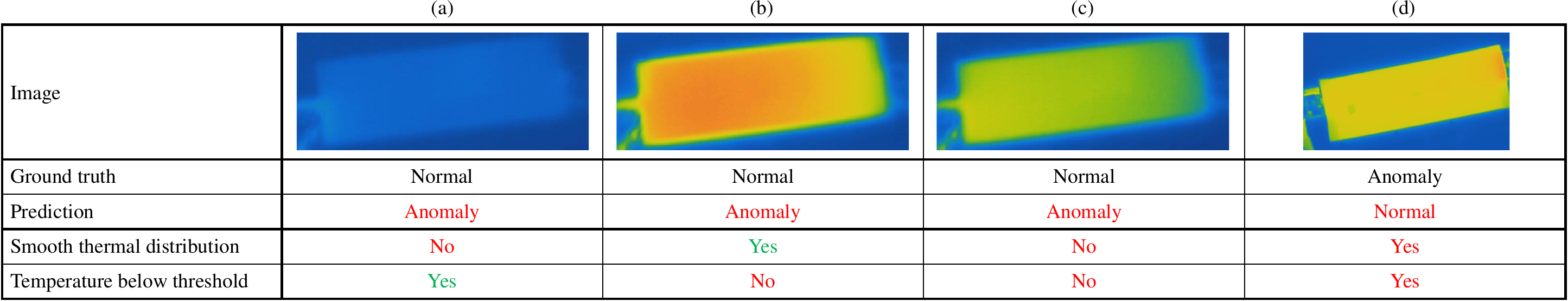}
  \vspace{-5mm}
  \caption{Samples of incorrect predictions in ChatGPT-4o, with the detected condition from the prior knowledge indicated in its explanation. Red text represents incorrect predictions, while green text represents correct predictions.}
  \vspace{-5mm}
  \label{fig:failure_case}
\end{figure*}

\subsection{Results}
\label{subsec:results}
We discuss the quantitative and qualitative results of our experiments in this subsection. 

\subsubsection{Output examples} Figure~\ref{fig:output_examples} shows output examples from the three VQA models on normal and anomalous data. ChatGPT-4o and LLaVa-13b also provide explanations before making the final anomaly detection prediction. These explanations can be used to further analyze the model, for example, to understand the reasoning behind incorrect predictions.

\subsubsection{Accuracy across different models} Table~\ref{tab:results} presents the accuracy (\%) averaged across multiple trials for different models (Avg. Acc.). Overall, ChatGPT-4o achieves the highest accuracy among the tested VQA models. Averaging the Avg. Acc. across five different prompts on the entire test set, ChatGPT-4o, LLaVa-13b, and BLIP-2 achieve 77.66\%, 59.66\%, and 61.66\%, respectively. These results highlight the importance of selecting an appropriate VQA model.

\subsubsection{Accuracy across different prompts} 
Table~\ref{tab:results} shows that Avg. Acc. varies across five prompts, with a range (min-max) across prompts of 11.3\% (ChatGPT-4o), 16.7\% (LLaVa-13b), and 33.0\% (BLIP-2). This highlights the sensitivity of performance to prompt selection, making it a key hyperparameter in model selection.

\subsubsection{Accuracy on normal versus anomaly} 
As seen in Table~\ref{tab:results},  Avg. Acc. tends to be higher for anomalous data, suggesting a bias toward false positives. This can be caused by normal data must satisfy both prior knowledge conditions, so misclassification of either can lead to incorrect predictions (Figure~\ref{fig:failure_case}(a)-(b)). In Figure~\ref{fig:failure_case}(a), subtle color changes on the left part of the image may cause misclassification. In Figure~\ref{fig:failure_case}(b), a temperature close to 50°C leads to an overheating prediction. In addition to these cases, errors can also arise from misclassifications in both conditions (Figure~\ref{fig:failure_case}(c)).

\begin{figure}[]
  \centering
  \includegraphics[width=\linewidth]{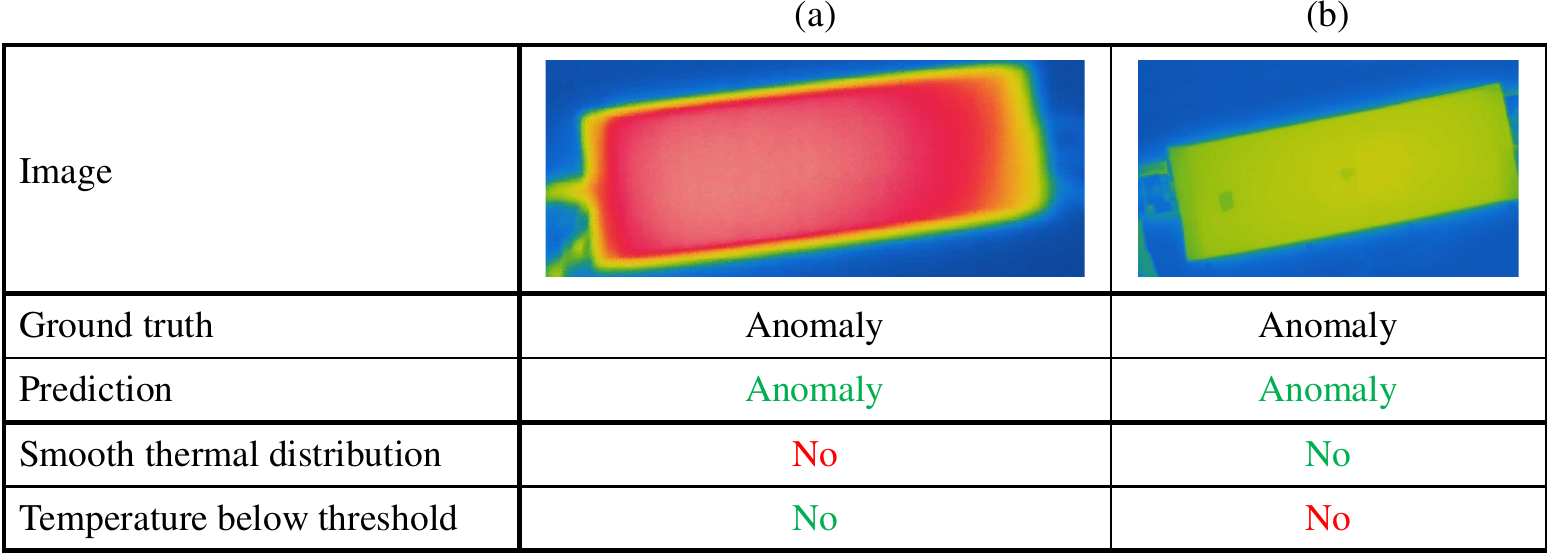}
  \vspace{-5mm}
  \caption{Samples of correct predictions in ChatGPT-4o, with incorrectly predicted conditions from the prior knowledge indicated in its explanation. Red text represents incorrect predictions, while green text represents correct predictions.}
  \vspace{-2mm}
  \label{fig:correct_but_wrongreasoning_case}
\end{figure}

Additionally, Figure~\ref{fig:correct_but_wrongreasoning_case} illustrates cases where reasoning errors still yield correct predictions on anomalous data, leading to higher accuracy on anomalous data. For example, in Figure~\ref{fig:correct_but_wrongreasoning_case}(a), an incorrect anomaly attribution does not affect the final outcome. Similarly, in Figure~\ref{fig:correct_but_wrongreasoning_case}(b), a mistaken high-temperature detection still leads to a correct prediction.

\subsubsection{Accuracy on different types of anomalous data} Comparing Avg. Acc. (overheating), Avg. Acc. (reflection), and Avg. Acc. (spatial tape) in Table~\ref{tab:results}, the reflection case appears to be the easiest to detect. This is because reflection anomalies exhibit both uneven thermal distribution and overheating conditions, making them more distinguishable compared to overheating and spatial tape cases, which involve only one condition. In particular, the uneven distribution in spatial tape cases can be very subtle, making it more difficult to detect, as seen in Figure~\ref{fig:failure_case}(d).

\subsubsection{Accuracy fluctuations across different trials} We report the range (max Acc. - min Acc.) across different trials in Table~\ref{tab:results}. Since accuracy varies between trials, conducting multiple trials helps achieve more reliable predictions and a better understanding of the model's consistency.

\subsubsection{Unsure cases} \label{subsubsec:unsure} 
\begin{figure}[]
  \centering
  \includegraphics[width=\linewidth]{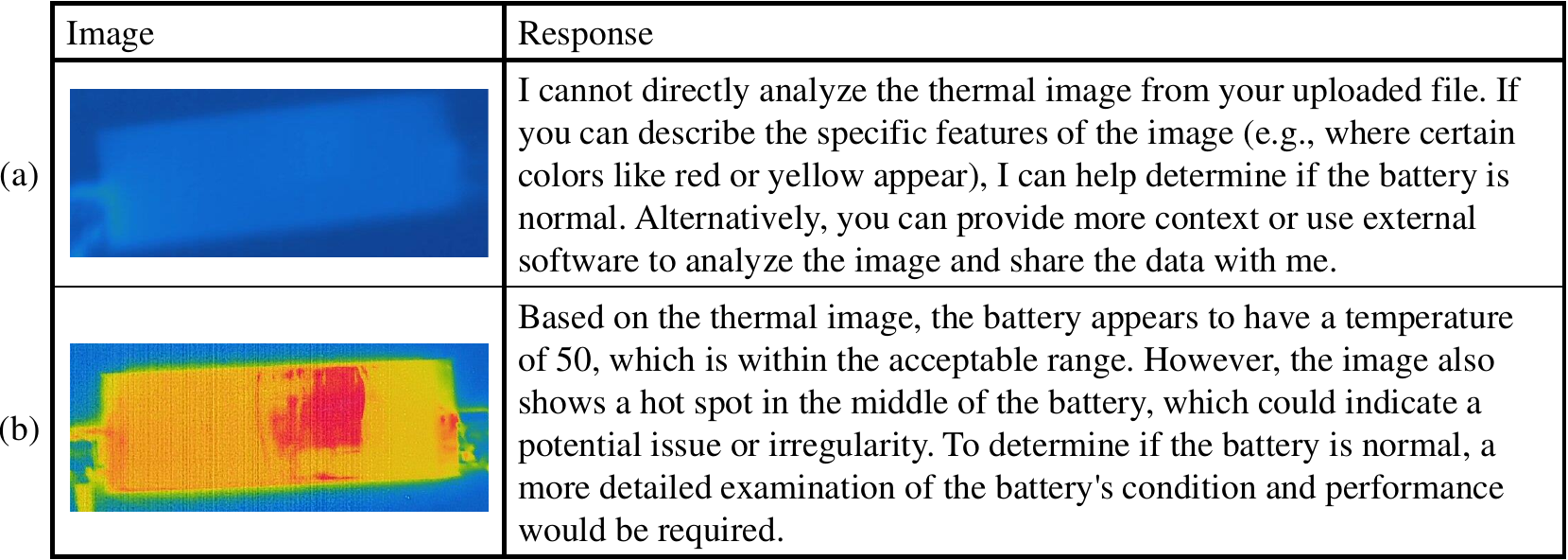}
  \vspace{-5mm}
  \caption{Samples of unsure predictions from the LLaVa-13b model on (a) normal and (b) anomalous data.}
  \vspace{-2mm}
  \label{fig:unsure_case}
\end{figure}

Table~\ref{tab:results} reports the percentage of unsure predictions (\%Unsure) for each prompt and model, with most coming from LLaVa-13b. Figure~\ref{fig:unsure_case} illustrates cases where the model either fails to determine conditions (Figure~\ref{fig:unsure_case}(a)) or detects conditions but does not provide a final prediction (Figure~\ref{fig:unsure_case}(b)).

\subsubsection{Comparisons to SOTA} We compare our method with SOTA methods in Table~\ref{tab:sota}. Following \cite{shabayek2025ai}, we evaluate performance using the Area Under the ROC Curve (AUC (\%)). We report the AUC of ChatGPT-4o with Prompt 2, averaged across five trials. Despite not using any training data, our method remains competitive with SOTA approaches. Furthermore, since our method does not rely on training data, it is unaffected by noisy training data.

\begin{table}[]
\centering
\caption{AUC (\%) comparisons with SOTA methods trained on battery data, using either clean or noisy data. The SOTA method results are taken from \cite{shabayek2025ai}. Our method is zero-shot and therefore does not require any training data.}
\resizebox{\linewidth}{!}{
\begin{tabular}{|l|c|c||l|c|c|}
\hline
Method      & \Centerstack{AUC (\%) \\ clean train}      & \Centerstack{AUC (\%) \\ noisy train}  & Method      &\Centerstack{AUC (\%) \\ clean train}      & \Centerstack{AUC (\%) \\ noisy train}   \\ \hline
CFLOW-AD \cite{gudovskiy2022cflow}    & 87.3 & 76.9  & STFPM   \cite{wang2021student}    & 96.1  & 87.1\\
PatchCore \cite{roth2022towards}  & 99.0 & 77.4 & CFA    \cite{lee2022cfa}     & 94.2  & 99.0 \\
FastFlow  \cite{yu2021fastflow}  & 100.0  & 79.3 & DRAEM   \cite{zavrtanik2021reconstruction}    & 99.1  & 92.2 \\
DFM  \cite{ahuja2019probabilistic}   & 99.6   & 80.3 & SimpleNet \cite{liu2023simplenet}  & 100.0 & 97.7 \\
EfficientAD \cite{batzner2024efficientad} & 100.0 & 81.0  & FAUAD  \cite{shabayek2025ai}     & 100.0  & 99.0 \\ \cline{4-6}
PaDiM       \cite{defard2021padim} & 99.6  & 86.3  & Ours (zero-shot)        & \multicolumn{2}{c|}{86.6}    \\ \hline
\end{tabular}
}
\vspace{-2mm}
\label{tab:sota}
\end{table}

\begin{figure*}
  \begin{minipage}{.3\linewidth}
    \centering
    \includegraphics[width=\linewidth]{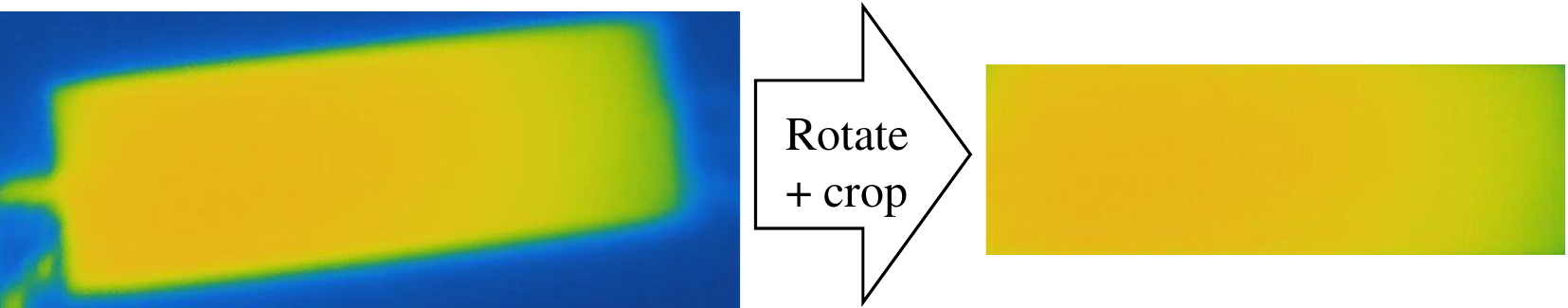}
    \caption{To remove background noise and enhance performance, especially for normal data, the test data is cropped and rotated.}
    \label{fig:preprocessing}
  \end{minipage}
  \hfill
  \begin{minipage}{.7\linewidth}
    \centering
    
    \captionof{table}{Comparison of Avg. Acc. (\%) before and after pre-processing. We focus on models and prompts that underperformed on normal data in Table \ref{tab:results}. The better performance between before and after pre-processing is marked in bold. BLIP-2 performance remains unchanged.}
    \resizebox{\linewidth}{!}{
    \begin{tabular}{|l||cccccc|cccccc|}
\hline
                    & \multicolumn{6}{c|}{Before pre-processing}                                                                               & \multicolumn{6}{c|}{After pre-proprecessing}                                                                               \\ \hline
Model               & \multicolumn{2}{c|}{ChatGPT-4o}                    & \multicolumn{4}{c|}{LLaVa-13b}                                & \multicolumn{2}{c|}{ChatGPT-4o}                    & \multicolumn{4}{c|}{LLaVa-13b}                               \\ \hline
Prompt              & 1             & \multicolumn{1}{c|}{3}             & 1             & 2             & 4             & 5             & 1             & \multicolumn{1}{c|}{3}             & 1             & 2             & 4             & 5            \\ \hline \hline
Avg. Acc. (all)     & 73.0          & \multicolumn{1}{c|}{71.0}          & 58.3          & 55.6          & 63.3          & \textbf{52.2} & \textbf{83.0} & \multicolumn{1}{c|}{\textbf{82.3}} & \textbf{63.9} & \textbf{56.7} & \textbf{68.3} & 51.1         \\ \hline
Avg. Acc. (normal)  & 42.2          & \multicolumn{1}{c|}{37.8}          & 11.1          & 2.5           & 29.6          & \textbf{6.2}  & \textbf{65.2} & \multicolumn{1}{c|}{\textbf{63.7}} & \textbf{49.4} & \textbf{7.4}  & \textbf{50.6} & \textbf{6.2} \\
Avg. Acc. (anomaly) & \textbf{98.2} & \multicolumn{1}{c|}{\textbf{98.2}} & \textbf{97.0} & \textbf{99.0} & \textbf{90.9} & \textbf{89.9} & 97.6          & \multicolumn{1}{c|}{97.6}          & 75.8          & 97.0          & 82.8          & 87.9         \\ \hline
\end{tabular}
    }
    \label{tab:preprocessing}
  \end{minipage}
  \vspace{-2mm}
\end{figure*}

\subsubsection{Increasing performance on normal data} One weakness of this method is its performance on normal data. We believe that the noisy background may contribute to an uneven thermal distribution. Therefore, we apply rotation and cropping to the test data to remove the background, as illustrated in Figure~\ref{fig:preprocessing}. The results (Table~\ref{tab:preprocessing}) show that in some cases, preprocessing significantly improves performance on normal data. Although performance on anomalous data slightly declines, the substantial improvement in normal data leads to an overall performance increase. Thus, preprocessing thermal images is one of the way to enhance performance.

\section{Conclusion and Future Work}
\label{sec:conclusion}
We explore the possibility of zero-shot learning for anomaly detection in battery thermal images using VQA with prior knowledge of the normal data. Our findings suggest that this approach is promising despite its simplicity as it can compete with SOTA methods even without any training data. However, we identify several weaknesses. First, the method struggles with normal data classification. One simple solution is to preprocess the image to remove unnecessary background. Future work can also incorporate one or a few normal data samples as visual context, making it a one-shot or few-shot approach. Second, predictions can fluctuate across trials, so performing multiple trials is recommended for more reliable results. Third, different prompts can yield varying outcomes, making it necessary to identify the best prompt for each model.

\bibliographystyle{IEEEtran}
\bibliography{submission}

\end{document}